\documentclass[runningheads]{llncs}

\usepackage[T1]{fontenc}
\usepackage{graphicx}
\usepackage{enumitem}
\usepackage{cite}
\usepackage{algorithmic}
\usepackage{graphicx}
\usepackage{textcomp}
\usepackage{comment}
\usepackage{hyperref}
\usepackage{tabularx}
\usepackage{microtype}
\usepackage{pifont}
\usepackage{listings}
\usepackage[table]{xcolor}
\usepackage{multirow}
\usepackage[most]{tcolorbox}
\usepackage{array}
\usepackage{pifont}
\usepackage{amssymb,amsfonts}

\newboolean{showcomments}
\setboolean{showcomments}{true} 
\ifthenelse{\boolean{showcomments}}
  {
		\newcommand{\nbb}[2]{
		\fcolorbox{black}{yellow}{\bfseries\sffamily\scriptsize#1}
		{\sf$\blacktriangleright$\textcolor{blue}{\textit{#2}}$\blacktriangleleft$}
		}
		
		\newcommand{\remarks}[1]{\color{red}[#1]\color{black}}

		\newcommand{\del}[1]{\textcolor{red}{\sout{#1}}} 
  }
  {
		\newcommand{\nbb}[2]{}
		\newcommand{\remarks}[1]{}

		\newcommand{\del}[1]{} 
  }

\begin{document}

\newtcolorbox{boxA}{
    fontupper = \bf,
    boxrule = 1.5pt,
    colframe = black, 
    coltitle=black
}
\newtcolorbox{boxB}{
    fontupper = \it,
    boxrule = 1.5pt,
    colframe = black, 
    coltitle=black
}

\title{Towards Systematic Qualification of Vision-Language Models for Automotive Perception Systems}

\titlerunning{A qualification workflow for VLMs for Traffic Understanding}

\author{Malsha Ashani Mahawatta Dona \orcidID{0009-0003-3812-5466} \and
Konstantinos Rokanas\and
Alexander Säfström \and
Krishna Ronanki \orcidID{0009-0001-8242-6771} \and
Christian Berger\orcidID{0000-0002-4828-1150}}
\authorrunning{M.A.M.~Dona et al.}

\institute{
University of Gothenburg and Chalmers University of Technology, Gothenburg, Sweden
\email{\{malsha.mahawatta, krishna.ronanki, christian.berger\}@gu.se}\\
}

\maketitle              

\begingroup
\renewcommand{\thefootnote}{}
\footnotetext{\scriptsize
\textbf{Declaration} According to the Springer AI risk-assessment framework for responsible AI use in research, the authors declare that AI was used to polish and refine the language of this manuscript. The authors take full responsibility for the final content.
}
\addtocounter{footnote}{-1}
\endgroup

\begin{abstract}
The field of Artificial Intelligence (AI) is vastly growing and has been adopted for many application domains. Vision Language Models (VLMs) are one of the recently advanced AI techniques that have been explored to support automotive features such as vehicle perception, planning, and safety assurance. However, such language models are prone to hallucinations, posing a potential threat to the safety of automotive systems that may incorporate them. Within the automotive domain, VLMs could not only hallucinate traffic objects, but could also fail to identify traffic objects that are actually present, which may potentially lead to dangerous situations. Though we have observed a growing body of literature that proposes verification and validation (V\&V) techniques for safe and trustworthy AI, these methods are often studied in isolation, focusing either on run-time or design-time phases. Such isolated techniques could be insufficient in safety-critical, realistic contexts such as automotive perception systems. In this paper, we analyze both design-time and run-time verification and validation techniques based on a taxonomy of trustworthiness presented by Huang et al. We present an automotive study in which a design-time qualification workflow is proposed to complement run-time monitoring. This workflow combines a fixed safety-relevant ontology-based structured annotation system together with a synonym-based evaluation process to statistically evaluate three state-of-the-art VLMs against data from the nuScenes dataset. We observed that the proposed technique enables deterministic and repeatable quantification of the hallucinations VLMs generate in automotive perception-related tasks. The proposed workflow supports model comparison and deployment-oriented engineering decisions within the design-time verification and validation process and will contribute to a holistic verification strategy that strives towards trustworthy automotive perception systems.

\end{abstract}

\section{Introduction}
\label{sec:intro}

Modern software systems have benefited from a combination of Artificial Intelligence (AI), Computer Vision (CV), and sensor fusion, which could potentially lead to enhanced 
decision-making, perception, and efficiency. The recent developments in the automotive industry increasingly integrate AI components into perception monitoring systems, decision making systems and other driver assistance functionalities enhancing traffic safety and traffic efficiency \cite{li2025applications}. This technological shift tends to embrace end-to-end approaches that include algorithm frameworks, utilizing raw sensor inputs \cite{10614862}. Among many other relevant AI techniques, the usage of LLMs for real-time processing of sensor input such as camera footage, LiDAR, and radar in self-driving vehicles has recently gained the attention of the research community in the automotive context  \cite{rony2023carexpertleveraginglargelanguage}. Vision Language Models (VLMs) are a type of LLMs that work with such multimodal data in various domains that have shown exemplary capabilities in visual input-related tasks, not only limited to image captioning, multi-modal reasoning, and visual question answering \cite{dona2024llms}. Henceforth, VLMs are now widely assessed to explore their potential application in automotive contexts to support various tasks such as perception and monitoring tasks, and in-car passenger communication \cite{hybridReasonongLLMinCars}.

\subsection{Problem Domain and Motivation}
\label{sec:ProblemAndMotivation}
In safety-critical domains, failures of AI components would pose system-level hazards. Though AI components such as VLMs have shown remarkable capabilities in automotive domain-related tasks, the potential risk of hallucinations generated by such models has remained an open challenge. Hallucinations are essentially known as nonsensical or unfaithful information that are generated by LLMs compared to the provided context or real-world knowledge \cite{huang2023survey_hallucinations}. Within the automotive context, three types of hallucinations have been mainly identified; where the answer generated by a VLM refers to objects that are not present in the actual context, where the VLM fails to identify objects that are actually present in the context, and instances where the VLM rejects processing the images due to conflicting reasons that are beyond the understanding of the authors \cite{dona2024evaluating}. From the perspective of a software engineer, the stochastic behavior of VLMs that may produce incorrect outputs would challenge traditional verification, testing, and qualification methods\cite{10.1016/j.jss.2021.111050}. 


Recent research has shown interest in detecting and mitigating hallucinations in various contexts such as code generation, question and answer scenarios, and automotive perception tasks \cite{ manakul2023selfcheckgptzeroresourceblackboxhallucination,dona2024llms}. Among these methods, the use of LLMs/VLMs as a judge to verify their own outputs \cite{manakul2023selfcheckgptzeroresourceblackboxhallucination} has shown promising potential in detecting hallucinations in real-time evaluations where the ground truth is not present. Such techniques can be considered as run-time-oriented verification techniques \cite{Huang2024}, in which decisions can be made without relying on the ground truth by principle.

While such approaches are valuable for detecting failures during operations, the LLMs/VLMs could also occasionally hallucinate when used as judges. Therefore, it is important to introduce a qualification step to guide early-phase engineering decisions and to evaluate existing LLMs/VLMs, with the focus on avoiding the recursive problem of hallucinations \cite{NEURIPS2023_91f18a12}. Such a qualification step would especially be important within safety-critical systems, where architectural and model selection decisions must be made prior to deployment. 

\subsection{Research Goal and Research Questions}
\label{sec:RG_RQs}
Recent taxonomies of the safety and trustworthiness of LLMs establish fundamental differences between design-time (offline) and run-time (online) evaluation techniques \cite{Huang2024}, emphasizing that both techniques are necessary to develop trustworthy AI systems. The design-time evaluation techniques are capable of assessing AI components prior to deployment, supporting model selection decisions, system integration, and risk assessment, whereas run-time evaluation techniques focus on detecting and mitigating potential failures occurring during system operations. 

In this study, we focus on a design-time verification technique and establish a controlled evaluation framework that follows a statistical approach to systematically evaluate VLMs for potential use within a safety-critical systems. We follow the taxonomy presented by Huang et al.~\cite{Huang2024} to discuss existing evaluation techniques that fall under the design-time phase and safeguarding techniques that fall under the run-time monitoring phase. For the run-time evaluation, we implement an existing ``LLM-as-a-judge''\cite{NEURIPS2023_91f18a12} verification technique that assess the AI system during run-time, complementing the design-time verification process.  

\begin{enumerate}[leftmargin=*, label={\textbf{RQ-\arabic*}}]
    \item How can design-time verification techniques be used to systematically evaluate hallucination risks of VLMs as a qualification step before deploying them in safety-critical systems to support perception-related tasks? 
    \item How do existing verification and validation techniques differ in their applicability and limitations when used at design-time compared to run-time in automotive perception systems?
    \item What are strengths and limitations of run-time hallucination techniques when considered for complementing design-time verifications?
    \item To what extent can a statistically grounded design-time evaluation framework support engineering decisions related to model selection, qualification, and deployment readiness in safety-critical automotive perception tasks?
\end{enumerate}

\subsection{Contributions and Scope}
\label{sec:Contributions}

We discuss existing verification and validation techniques that enable the trustworthiness of an LLM-assisted perception system in both design-time and run-time operations based on the taxonomy proposed by Huang et al.~\cite{Huang2024}. We further explore the hallucination detection methods that use LLMs/VLMs to check their own results following the ``LLM-as-a-Judge'' technique \cite{dona2024llms}. The main contribution of this study is the design-time evaluation approach that can be suggested as a qualification step in the process of using LLMs/VLMs within perception monitoring systems for a safety-critical system, for instance, within vehicles. We further investigate how the ethical usage of AI components and falsification and evaluation techniques would complement each other, striving for the trustworthiness of the AI-based software systems. The replication package, including the supplementary materials used for this study, is available on \href{https://github.com/MalshaMahawatta/ICTSS2026-Systematic-Qualification-of-Vision-Language-Models-for-Automotive-Perception-Systems.git}{Github}.

\subsection{Structure of the Paper}
\label{sec:StructureOfThePaper}

The rest of the paper is organized as follows: Section \ref{sec:relatedWork} reviews the existing literature related to hallucination detection techniques whereas Section \ref{sec:methodology} explains the experiment pipeline and detailed method used for this study. Section \ref{sec:results} presents the results of the experiments, Section \ref{sec:Analysis_Discussion} presents the analysis and discussion of the results, and Section \ref{sec:conclusion_futureWork} concludes the paper.  

\section{Related Work}
\label{sec:relatedWork}

Verification and validation of AI-based complex software systems have become increasingly important with the adoption of LLM-powered modules. We often see non-deterministic behaviors of such LLM-based modules that challenge the traditional Verification and Validation (V\&V) practices \cite{11127266}. In safety-critical domains such as the automotive industry, standard guidelines, for instance, ISO/PAS 21448 and  ISO 26262 are in place to emphasize systematic qualification and risk reduction \cite{Kirovskii_2019}. However, modern Advanced Driver Assistance Systems (ADAS) and Autonomous Driving (AD) features that utilize VLMs/LLMs fall outside the scope of such standards and guidelines due to their limited concrete guidance on the operationalization of V\&V processes \cite{salay2017analysis}. Therefore, the majority of the existing V\&V techniques are largely isolated in engineering workflows in terms of design-time and run-time stages. 

In the related literature, studies showcase various falsification techniques and evaluation tactics for design-time phases of the LLM life cycle. Techniques such as prompt injection can challenge the behavior of an LLM-based system by injecting malicious instructions into the prompt \cite{perez2022ignore}. Such studies uncover to what extent LLM-based systems can be manipulated during design time. However, there seems to be no universal solution to these issues, for instance, a single methodology or an evaluation protocol that is sufficiently reliable in detecting such manipulations across different models and different domains that utilize multi-modal data.

As a V\&V  technique, studies that compare AI output with human experts report promising results in evaluating AI modules during their design time \cite{guo2023close}. Most studies exclusively focus on domains that involve only text-based data, narrowing their applicability in automotive industry perception monitoring systems that largely depend on multi-modal data. The most recent studies, such as \cite{qi2025safety}, establish a useful foundation in the field of safety assessment for the automotive industry in the age of AI. In addition to that, techniques such as benchmarks \cite{chen2024chatgpt} and statistical evaluation methods \cite{sun2018testing} are also prominent in V\&V processes during design-time stages; however, they are not investigated sufficiently to draw conclusions about their applicability in the automotive context.

``LLM-as-a-judge'' techniques have been deemed useful in evaluating LLM/VLM-assisted systems that often apply to run-time environments where the decisions have to be made without having access to ground truth data. The related literature suggests many LLM-based evaluation methods of different functionalities, methodologies, and application domains \cite{dona2024llms}. Some of such LLM-based evaluation techniques utilize single models \cite{fu2023gptscore}, whereas some utilize multi-model architectures \cite{manakul2023selfcheckgptzeroresourceblackboxhallucination}. However, the majority of these ``LLM-as-a-judge'' techniques aim at text-based inputs that may or may not be LLM-generated responses. 

The SelfCheckGPT technique presented by Manakul et al.~\cite{manakul2023selfcheckgptzeroresourceblackboxhallucination} includes an LLM-based hallucination detection technique that can be used to check LLM-generated responses. The LLM responses are generated multiple times for the same prompt and later used as context to check the consistency of the first response by decomposing it into single sentences. The consistency checking is conducted using LLMs by prompting them with each sentence together with one of the remaining responses to evaluate whether this sentence is supported by the response. The binary answers received for each sentence are recorded to calculate the sentence-wise consistency score. This technique has shown promising results when evaluated against the Wikibio dataset curated by the authors; however, its applicability in other application contexts, for instance, when multi-modal data is involved, has not been addressed. 

Sawczyn et al.~present a hallucination detection technique named FactSelfCheck \cite{sawczyn2025factselfcheckfactlevelblackboxhallucination} that uses the ``LLM-as-a-judge'' concept. This technique performs fact-checking using a different LLM to directly test the fact-level information that was in the response. It does not include any sentence-decomposing tasks, but rather the entire fact is represented as a knowledge graph, which is then verified by another LLM. While this method can be adopted to work with multi-modal data, it would not be efficient within a real-time environment, such as automotive, where having an intermediary layer that maps visual data into knowledge triplets would slow down the overall process. In addition to that, having only LLMs as evaluators would introduce the recursive problem of hallucinations as well. 

Dona et al.~\cite{dona2024llms} presented an adaptation of SelfCheckGPT that is applicable within the automotive domain to detect hallucinations in perception-related tasks when LLMs are used to detect objects. This approach uses multi-modal data and has been influenced by the CrossCheckGPT technique \cite{sun2024crosscheckgpt}. In this technique, the authors have proposed an exclusion criterion, where less consistent sentences, ie., hallucinated content, will be removed from the first response. This method has been evaluated with two state-of-the-art automotive datasets, focusing on further external factors such as weather conditions, lighting conditions, or driving patterns. This technique caters to the automotive domain, detecting and mitigating hallucinations in LLM-assisted perception and monitoring systems. 

\section{Research Methodology}
\label{sec:methodology}
We adopt an empirical methodology to investigate how design-time verification techniques complement run-time monitoring techniques to detect and mitigate hallucinations on the example of an automotive perception system. A taxonomy-guided V\&V categorization technique is first used to systematically evaluate the existing V\&V techniques that assess the trustworthiness of LLMs within safety-critical systems. This study considers hallucination detection and mitigation as quality assurance and verification processes that span across the run-time and design-time phases of the software engineering life cycle.  

\subsection{Taxonomy-based study selection}

\begin{figure*}
\centering
    \includegraphics[width=1\linewidth]{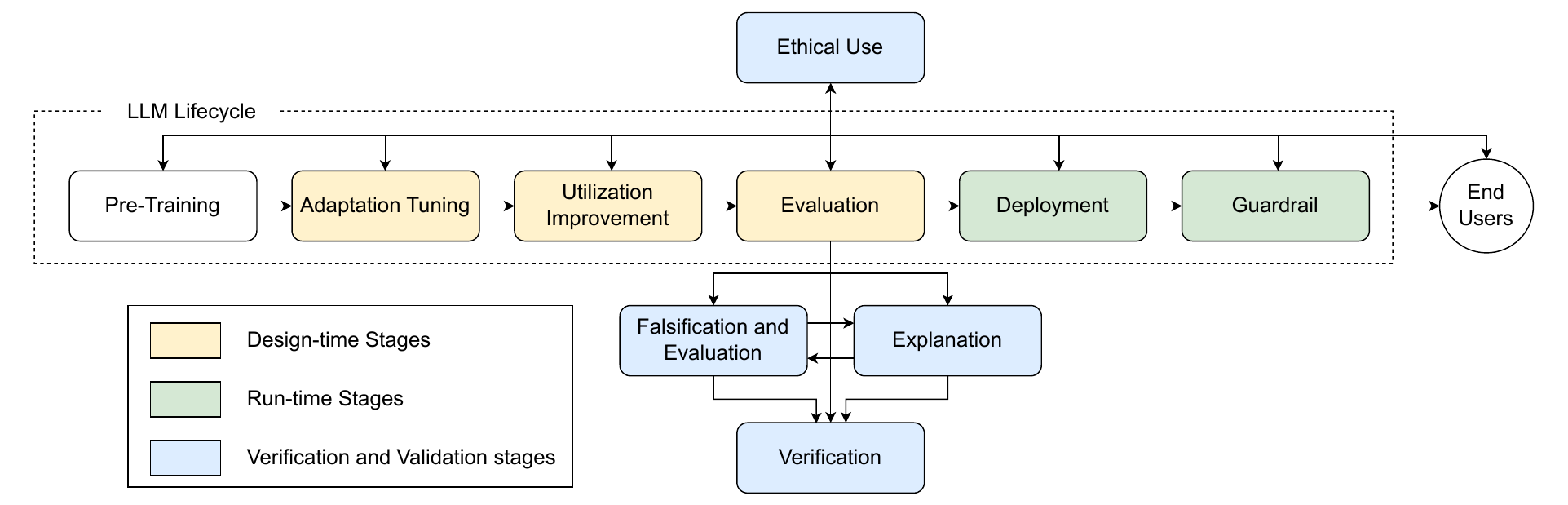}
    \caption{Verification framework in LLM lifecycle (Image: \cite{Huang2024})}
    \label{fig:VVFramework}
\end{figure*}

\begin{figure*}
\centering
    \includegraphics[width=1\linewidth]{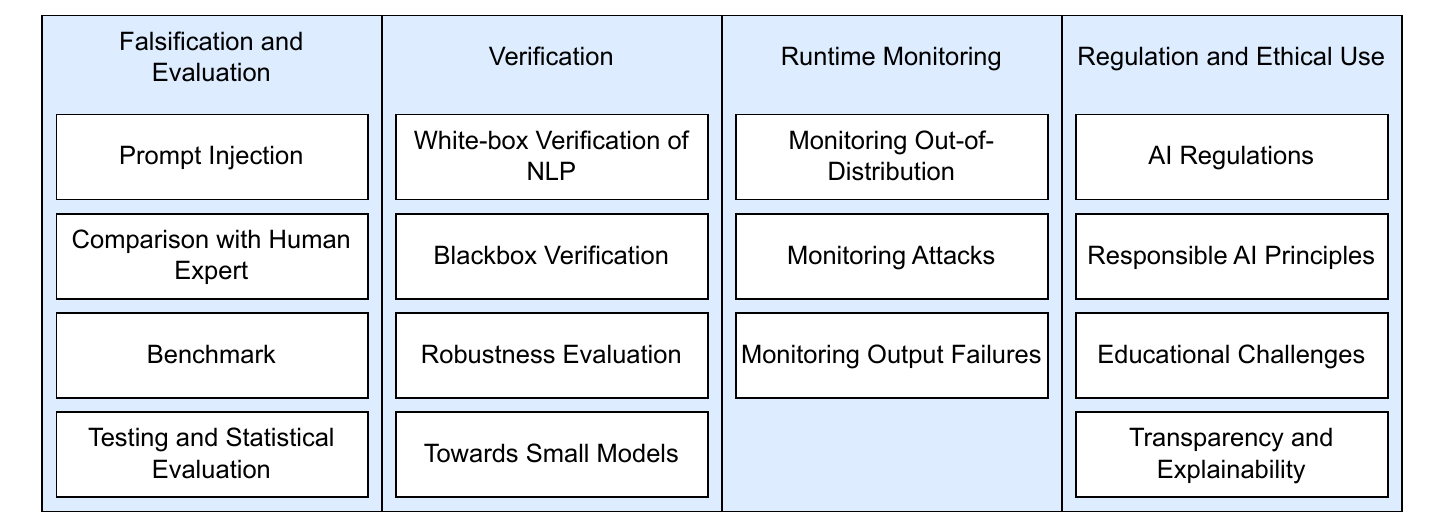}
    \caption{The Verification and Validation technique categorization based on the taxonomy presented by Huang et al. (Image: \cite{Huang2024})}
    \label{fig:VVCategorization}
\end{figure*}

Figure \ref{fig:VVFramework} shows the taxonomy proposed by Huang et al.~\cite{Huang2024} categorizing V\&V techniques onto the LLM lifecycle. For this study, we selectively focused on the categories Falsification and Evaluation, Verification, Run-time Monitoring, and Regulations and Ethical Use, and the respective V\&V techniques that fall under them (cf.~Figure.~\ref{fig:VVCategorization}). The studies presented in \cite{Huang2024} have been categorized under the taxonomy below and were analyzed further to understand their application domain, modality of the data, and limitations. The following sections reproduce and summarize essential parts from the study by Huang et al.~\cite{Huang2024}.

\subsubsection{Falsification and Evaluation}
The verification techniques that identify and evaluate the scenarios where LLMs actively fail, posing benign or malignant risks, fall under this category \cite{Huang2024}. The prompt injection techniques \cite{perez2022ignore}, comparisons with human experts \cite{guo2023close}, evaluations against benchmark datasets \cite{wang2023robustness}, and testing and statistical evaluations are some V\&V techniques that are identified. 

\subsubsection{Verification}
The techniques that evaluate LLMs based on Natural Language Processing (NLP) approaches are categorized under this category. Verification based on NLP models \cite{gowal2018effectiveness}, blackbox verification \cite{wicker2018feature}, robustness evaluation on LLMs, and use of smaller models are some relevant V\&V techniques. 

\subsubsection{Run-time Monitoring}

The run-time evaluation processes provide monitoring techniques to secure LLMs while they interact with the end users \cite{Huang2024}. Monitoring Out-of-Distribution (ODD) scenarios, attacks, and output failures are some common V\&V techniques that can be considered as runtime monitoring \cite{Huang2024}.

\subsubsection{Regulations and Ethical Use}
Technical means for run-time and design-time monitoring are becoming insufficient to verify and validate ever-growing, AI-based, complex systems. The technical aspects must be complemented by ethical means to ensure that AI-based systems align with human principles and values. Decisions related to regulations, implementation of AI acts, Agreements on responsible AI principles, and Educational Challenges are some of the aspects that need to be covered to regulate complex AI systems while ensuring ethical usage \cite{altai, Huang2024}.

Based on the above taxonomy, we select two complementary classes for our study: (a) design-time statistical evaluation and (b) run-time self-checking approaches covering the engineering lifecycle of LLM-based complex systems. 

\begin{figure*}
\centering
    \includegraphics[width=1\linewidth]{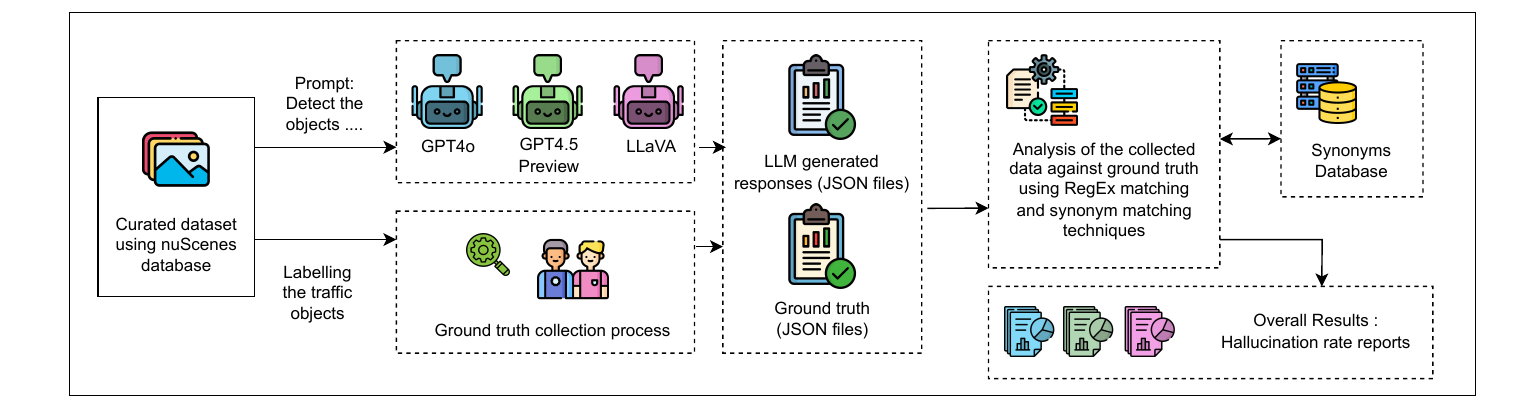}
    \caption{Overview diagram of the experimental setup: All curated images are passed to a set of VLMs (GPT4o and GPT4.5-preview, and LLaVA) with a prompt that guides the VLMs to detect the traffic entities that are visible. The same image set is manually evaluated by a group of human annotators to record the ground truth, ie., accurate object labels. The collected ground truth and the VLM-generated responses are statistically evaluated to analyze the performance of the VLMs. (Icons: Flaticon.com)}
    \label{fig:experimentPipeline}
\end{figure*}

\subsection{Design-Time Statistical Evaluation}

The proposed evaluation technique strives to assess a model's proneness to hallucinate before the deployment of the AI component. The VLMs are considered black boxes, and their behavioral stability and the consistency of the outputs are recorded for perception monitoring tasks. The experiment pipeline is outlined in Figure \ref{fig:experimentPipeline}. We curated an image set using nuScenes \cite{nuscenes}, a widely accepted dataset in the automotive context due to its realistic and comprehensive multi-modal image and sensor data. The selected images were used as input to the three VLMs (GPT4o and GPT4.5-preview, and LLaVA (llava:latest 8dd30f6b0cb1)) with a prompt that included predefined rules. VLM responses were recorded and statistically analyzed against the ground truth to answer the research questions. 

\subsubsection{Dataset Curation}

nuScenes \cite{nuscenes} is a widely adopted benchmark dataset that provides diverse urban driving scenarios curated for automotive domain-related research. This dataset contains 1000 scenes that were collected in Boston, USA, and Singapore, focusing on diverse locations and driving patterns. Each scene is 20s long and covers high traffic density, rare classes, and maneuvers \cite{nuscenes}. The dataset provides annotations for 23 classes. 

For our experiment, we curated a subset of 1000 images from nuScenes, focusing on right-hand traffic scenes collected in Boston. The selected subset of images contained images captured during the day by the vehicle front camera. Given the annotation inaccuracies reported in relevant literature \cite{Chan_Li_Baris_Sadiq_Donzella_2024,nuscenes_devkit_issue366}, the curated dataset was manually annotated to record object labels as ground truth. 

\subsubsection{Data Collection}

As depicted in the experimental pipeline (cf.~Figure \ref{fig:experimentPipeline}), the curated subset of data was fed into three VLMs with a pre-defined prompt. We used two proprietary VLMs: GPT4o \cite{openai2024gpt4ocard} and GPT4.5-preview \cite{openai2024gpt45}, and one locally executable open VLM, Large Language-and-Vision Assistant (LLaVA)~\cite{liu2023visualinstructiontuning}. Each VLM was tested with the following prompt to identify the objects in each image. 

\begin{verbatim}
You are analyzing a traffic scene image for automotive perception.
Predefined relevant objects: car, bus, truck, bicycle, motorcycle, 
pedestrian, traffic light (red, green, yellow).
\end{verbatim}

We focus on the most common and most relevant objects for traffic safety by enforcing a set of rules as a guide for the VLM. The annotators also followed the same set of rules during the manual annotation process to generate ground truth object labels. The VLM responses for each image were recorded and later processed with an automated text-matching process against the ground truth to evaluate the VLM performance. 

\begin{boxB}
\noindent \textbf{Rule 1} Only detect objects that are at the same level as the vehicle taking the photo (objects on the road). 

\noindent \textbf{Rule 2} Do NOT include objects or pedestrians on sidewalks.

\noindent \textbf{Rule 3} The ONLY exception is traffic lights, which should always be detected regardless of position as long as you can distinguish their color.

\noindent \textbf{Rule 4} Return a list of each object from the list of predefined relevant objects that you see in this image.

\noindent \textbf{Rule 5} If you identify multiple objects of the same type, you must output their occurrence an equal number of times to the observed cardinality.

\noindent \textbf{Rule 6} In the case that you do not see any relevant objects, simply return an empty output.

\noindent \textbf{Rule 7} Your response must be a comma-separated list of detected objects, with no additional text, formatting, or code blocks. Example: car, car, traffic light red. 

\noindent \textbf{Rule 8} Be precise and thorough in your detection. Do not include objects that aren't clearly visible.
\end{boxB}

A custom tool was created to manually annotate the objects that are visible in each image. The objects visible were structured and used as ground truth during the VLM response evaluation phase. The annotation process was conducted by two human annotators, who independently worked on half of the dataset each. The annotators cross-validated each other's labels as the second step, where the inter-rater agreement was calculated based on Cohen's kappa \cite{McHugh2012kappa}. We report a value of 1 for Cohen's kappa, indicating high reliability levels between the decisions made by two annotators. 

\subsubsection{Data Analysis}
\label{Sec:DataAnalysis}

A list of synonyms was created using the WordNet synonym database \cite{wordnet2010} to treat different words that carry similar meanings. During the data analysis process, the VLM responses and ground truth labels were treated as multisets rather than sequences, and they were quantitatively compared using regular expressions to identify overlaps and deviations. 

In this study, we formulated the null hypothesis $H_0$ as there being no systematic difference in hallucination generation among the models, while the alternative hypothesis $H_1$ states that at least one of the models indicates a systematically different hallucination performance. We define hallucinations (h) as instances where VLM outputs (L) do not perfectly align with ground truth (G).  

\begin{equation}
\label{eqn:hallucinationDefinition}
    h \Longleftrightarrow L \neq G
\end{equation}

We encountered three types of hallucinations as: VLM-generated objects not in the ground truth, Ground truth objects not in VLM-generated objects, and Incorrect frequencies of reported objects in VLM-generated objects.

The quantity of the hallucinations (H) can be derived as: 

\begin{equation}
\label{eqn:QuantityOfhallucinations}
    H=|L \backslash G|+|G \backslash L|
\end{equation}




The normalized hallucination ratio was calculated for each image individually, taking into account the proportions of the number of hallucinations (H) within the union of ground truth (G) and VLM-generated objects (L).  

\begin{equation}
\label{eqn:NormalizedhallucinationRate}
    Normalized Hallucination Rate  = H/|\,G \cup L|\, 
\end{equation}

Based on the taxonomy, first, we consider this method as a comparison with human experts, given the human involvement in generating the ground truth. We also consider this technique as a statistical evaluation, given that human expertise is not directly involved in the automated evaluation. 

\subsection{Run-time Monitoring as a Complementary Baseline}

As the last step of the analysis, a comparison was conducted between one of the LLM-as-a-judge techniques presented by Dona et al.~\cite{dona2024llms} to understand how the proposed methodology can be supplemented by run-time safeguarding techniques to improve the trustworthiness of VLMs in safety-critical perception tasks. 

We implemented the SelfCheckGPT adaptation presented by Dona et al.~\cite{dona2024llms} as a baseline, using the same subset of images that were retrieved from the nuScenes dataset. We used LLaVA as the captioner model and GPT4o  as the checker model. The captioner model was prompted with each image five times to record the traffic objects that are visible. The first response was passed to the checker LLM, with one other response to check whether the first response is supported by the context response. Similarly, the first response of each image was checked against all four remaining responses, and the results were recorded.   

The checker LLM responses were analyzed to determine the hallucinated responses based on the consistency level score.  The responses with an average consistency larger than 0.5 were considered as non-hallucinated responses. Though the original approach decomposes paragraphs into sentences for the consistency check, it was not directly applicable in this case, where we dealt with a list of objects in the majority of the responses. Following the metrics proposed by Dona et al.~\cite{dona2024llms}, we treat correct responses as the positive class, whereas the consistent SelfCheck judgments are treated as positive predictions. Therefore, we consider True Positives (TP) as non-hallucinated responses flagged as non-hallucinated, False Positives (FP) as hallucinated responses flagged as non-hallucinated, True Negatives (TN) as hallucinated responses flagged as hallucinated, and False Negatives (FN) as non-hallucinated responses flagged as hallucinated. 

\section{Results}
\label{sec:results}


We report the results of the experiment under three main categories as shown in Table \ref{Tab:LLMPerformance}. A perfect match percentage is when the VLM is able to identify all objects correctly without a single hallucination. The open model LLaVA reports 7.8\% perfect matches, whereas it is higher, respectively 22.90\% and 28.70\% for proprietary models GPT4o and GPT4.5-preview. However, the second category indicates relatively higher values for all three models, where the focus was on indicating the extent to which the VLM output contained hallucinations. 92.20\% of LLaVA generated responses contained at least one hallucination, whereas it was respectively 77.10\% and 71.30\% for GPT4o and GPT4.5-preview. The normalized hallucination ratio was calculated as defined in Equation \ref{eqn:NormalizedhallucinationRate}. LLaVA yielded a higher hallucination rate of 73.30\%, where GPT4o and GPT4.5-preview displayed normalized hallucination rates of 50.78\% and 41.03\%, respectively. The last two rows of Table \ref{Tab:LLMPerformance} group the deviations in VLM responses under two error categories. The first category includes the number of instances where the VLM response contained a hallucinated traffic object that was not present in the ground truth. The second category represents the number of VLM responses where traffic entities were overlooked, but were reported to be visible in the image according to the ground truth. 

\begin{table}[]
\centering
\caption{Performances of VLMs under three categories: (a) Perfect matches where VLM successfully identified all hallucinations, (b) hallucinated images where VLMs returned at least one hallucination, (c) normalized hallucination rate indicating the average deviations occurred in the VLM outputs compared to ground truth and the deviations in VLM responses grouped by error categories: (d) hallucinating objects and (e) Overlooking Objects }
\label{Tab:LLMPerformance}
\begin{tabular}{l|l|l|l|}
\cline{2-4}
                                                    & LLaVA   & GPT4o   & GPT-4.5 \\ \hline
\multicolumn{1}{|l|}{(a) Perfect Matches}               & 7.8\%   & 22.90\% & 28.70\% \\ \hline
\multicolumn{1}{|l|}{(b) Hallucinated images}           & 92.20\% & 77.10\% & 71.30\% \\ \hline
\multicolumn{1}{|l|}{(c) Normalized Hallucination Rate} & 73.30\% & 50.78\% & 41.03\% \\ \hline
\multicolumn{1}{|l|}{(d) Hallucinating Objects} & 1418 & 154 & 252 \\ \hline
\multicolumn{1}{|l|}{(e) Overlooking Objects} & 2862 & 2123 & 1591 \\ \hline
\end{tabular}
\end{table}

We conducted a Shapiro-Wilk test \cite{shapiro1965analysis} to determine whether the data samples collected were normally distributed or not. P-values that are lower than the significance level were recorded for all three models. 

A Friedman test \cite{friedman1937use} was conducted to understand whether the statistical difference is significant between the related groups (cf.~Table~\ref{Tab:FriedmanTest}. The inputs for this test were the image subset and a data matrix of 3 columns that included the results generated by the three models.

\begin{table}[]
\centering
\caption{The results of the Friedman test indicate the significance of the statistical difference of the non-normalized data.}
\label{Tab:FriedmanTest}
\begin{tabular}{|l|l|}
\hline
Chi Square  & 797.4314    \\ \hline
P-value     & $p-value \ll 1e-10$ \\ \hline
Effect size & 0.3987      \\ \hline
\end{tabular}
\end{table}

Kendall W effect size calculation test \cite{tomczak2014need} was conducted on the results collected from the Friedman test to quantify the magnitude of the observed differences. We recorded an effect size of 0.3987 from this calculation. 


Table \ref{tab:selfcheckGPTComparison} reports the results of the SelfCheckGPT adaptation proposed by Dona et al.\cite{dona2024llms} when tested against the image subset we curated. The entire experiment was executed twice to verify the stability and reliability of the results, and the results are reported in the second and third columns of the table.

\begin{table}[h]
\centering
\caption{Results of SelfCheckGPT adaptation when tested with the image subset we curated.}
\label{tab:selfcheckGPTComparison}
\begin{tabular}{l|l|l|}
\cline{2-3}
                                        & First execution & Second execution \\ \hline
\multicolumn{1}{|l|}{True Positives}    & 57              & 57               \\ \hline
\multicolumn{1}{|l|}{False Positives}   & 5               & 5                \\ \hline
\multicolumn{1}{|l|}{True Negatives}    & 305             & 311              \\ \hline
\multicolumn{1}{|l|}{False Negatives}   & 633             & 627              \\ \hline
\multicolumn{1}{|l|}{Precision}         & 91.94\%         & 91.94\%          \\ \hline
\multicolumn{1}{|l|}{Recall}            & 8.26\%          & 8.33\%           \\ \hline
\multicolumn{1}{|l|}{Specificity}       & 98.39\%         & 98.42\%          \\ \hline
\multicolumn{1}{|l|}{Balanced accuracy} & 53.32\%         & 53.38\%          \\ \hline
\end{tabular}
\end{table}

In addition to that, the average response time was recorded across a sample size of ten prompts for each VLM as six seconds. The measured response time indicates that the SelfCheck configuration is not yet ideal for time-critical control loop decision systems unless better-performing small local models are deployed within more powerful hardware infrastructure.  

 \section{Analysis and Discussion}
\label{sec:Analysis_Discussion}

We first addressed \textbf{RQ-1} by systematically evaluating VLM responses to detect hallucinations. The results show that the proprietary models GPT4o and GPT4.5-Preview were less hallucinating compared to the open model LLaVA. However, it was noted that all three models were considerably hallucinating, where most of the images resulted in receiving hallucinated responses. When error categories were analyzed, it was noted that all models tend to overlook objects rather than hallucinate new objects, which has a critical impact in automotive application scenarios. We conducted further statistical analysis to check whether the deviations in hallucination detection were not random. The results of the Shapiro-Wilk analysis revealed that the data we collected do not follow a Gaussian distribution. Hence, a  Friedman test was conducted to derive the statistical significance of the collected data. The lower Friedman p-values indicated that the changes between the models are statistically significant, reflecting that the performances across models are not random. This led to the rejection of the null hypothesis $H_0$. However, neither the magnitude of the significance nor the effect is implied by these statistical tests. Based on the further Kendall W effect size test we conducted, we could conclude that our experiment only has a moderate effect size, which could not have impacted the p-value much.

The statistical approach proposed in this study includes a synonym check and a comparison using regular expressions to evaluate the VLM results with the ground truth. Therefore, this technique can be used during the design time to verify and validate the LLMs/VLMs, assisting engineers to systematically identify better models for perception-related tasks within safety-critical systems and make informed integration decisions.

To answer \textbf{RQ-2}, we systematically analyzed the existing V\&V techniques under the taxonomy presented by Huang et al.~\cite{Huang2024}. Our findings confirm that the majority of the evaluation techniques are standalone in either design-time or run-time, addressing different validation needs. In addition to that, our analysis reveals that there are only a handful of V\&V techniques that focus on either safety-critical domains such as automotive applications or use multi-modal data, allowing validation of VLMs as well \cite{qi2025safety,chen2024chatgpt}. The limitations identified through these studies reflect that verification and validation of LLM-based systems should be approached through the holistic life cycle of the AI component, starting from the design time itself. More studies should focus on V\&V  techniques catered for automotive systems where pre-deployment evidence is critical to make well-informed decisions to better prepare for run-time monitoring.

We addressed \textbf{RQ-3} by implementing the SelfCheckGPT adaptation proposed by Dona et al.~\cite{dona2024llms}. The results show higher precision and specificity values, indicating fewer false positives (FP). The lower recall values are a result of higher false negatives, indicating that many non-hallucinated objects were reported as hallucinations. The balanced accuracy field takes both TP and TN into account and reports moderately positive results for the SelfCheckGPT adaptation technique. The method works at the expense of mislabeling many responses as hallucinations, which should be addressed in future studies. While the method demonstrates that it is in principle applicable for hallucination detection as an LLM-as-a-judge technique, which works during run-time, it is adamant that there should be a qualification step to assess the VLMs/LLMs rather than relying on a run-time evaluation technique. 

\textbf{RQ-4} Compared to the LLM-as-a-judge technique, the statistical approach showed higher performance in terms of accuracy. However, since these techniques are fundamentally applicable in different stages of the LLM-based system life-cycle, a verification and validation pipeline that consists of both techniques could enhance the trustworthiness of AI-enabled complex systems. By relying only on ``LLM-as-a-judge'' techniques, we may open a recursive problem of VLMs/LLMs hallucinating even when they are employed as a judge. While the significance of such hallucination detection techniques is quite notable, especially since they can be executed in real-time environments without relying on ground truth data, the value of statistically grounded evaluation techniques is also important in instances where ground truth is available to process. Therefore, applying a statistical approach as a design-time step could help the VLM/LLM safeguarding process significantly.

For VLM-based perception modules to be integrated as a safety component, they should comply with the EU AI Act's high‑risk classification rules (Art.~6)~\cite{AIA}. In that case, model providers must satisfy the Chapter III requirements, which are usually operationalized through documented evaluations and monitoring of the work products. Our evaluation framework for VLM-assisted vehicle perception acts as a controlled, statistically grounded falsification and verification step. This contribution is significant within the LLM/VLM lifecycle because, in the EU AI Act, these expectations are formalized through requirements on continuous risk management, dataset governance, technical documentation and logging, transparency to deployers, human oversight, accuracy/robustness, and post-market monitoring (Articles 9–15 and 72)~\cite{AIA}. Our framework's outcomes can help operationalize the relevant AI Act obligations in practice (evaluation protocols, quantified reliability bounds, and failure-mode taxonomies). This supports accountable model selection and informs deployment constraints and run-time monitoring. Moreover, the framework also operationalizes core Trustworthy AI requirements identified by the AI HLEG\cite{altai}, such as technical robustness and safety, transparency, accountability, and human agency, by converting ethical principles into measurable evaluation criteria.

We assessed the threats to validity based on Feldt and Magazinius \cite{RobertFeldt_ThreatsToValidity}. In the initial dataset curation process, we limited the number of images to 1000 to keep the processing time at a reasonable length. In addition to that, considering the prior work done by Dona et al.~\cite{dona2024llms} on lighting conditions on images, we limited the diversity of our dataset by selecting only the daytime-captured images, which do not include adverse weather conditions. This would have narrowed the scope of the current study. 

Prompt programming is an important step in supporting the VLM/LLMs in generating accurate and relevant outputs of high quality. Though the final prompt we used for the experiments was obtained heuristically, given the scope of the study, we did not include further experiments on various other prompting techniques. The results of the experiments could significantly vary depending on external factors such as different model architectures, training datasets, etc. We adopted three different off-the-shelf VLMs to increase the diversity of the experiment pipeline with the intention of reducing potential inhibitors for generalization. However, we acknowledge that the results may suffer in generalizability given the unknown nature of the training data these models have been exposed to.

\section{Conclusion and Future Work} 
\label{sec:conclusion_futureWork}

Modern VLMs have found their way into the automotive domain due to their exceptional capabilities in processing multi-modal data, which will potentially lead to improved ADAS functions for autonomous systems. In this work, we investigate the roles of design-time and run-time V\&V techniques that can be applied to safety-critical automotive perception systems. We systematically analyzed the performance of three VLMs (GPT4o, GPT4.5-preview, and LLaVA) to understand to what extent these models hallucinate when they are being used for traffic object identification tasks. Due to the non-deterministic behavior we observed in VLM-based perception systems, we put forward the argument that run-time or design-time hallucination mitigation techniques alone are not sufficient to guarantee the trustworthiness of the perception-based systems. 

In the proposed approach, we focus on a design-time statistical evaluation approach with a run-time monitoring approach that uses an ``LLM-as-a-judge'' technique to detect hallucinations. However, considering the run-time hallucination detection requirements that persist within vehicles when VLMs are used to extend or complement perception stacks in Advanced Driver Assistant Systems (ADAS), LLM-as-a-judge techniques would fit better, given that they work without being dependent on ground truth data to spot hallucinations. The proposed statistical approach is applicable as a design-time evaluation technique within VLM safeguarding processes, where the ground truth is available. The selection of better-fitting VLMs for perception tasks can be based on the proposed approach.

We observed that statistical approaches produce deterministic scores in evaluating VLM outputs. All models we used for our experiments hallucinated quite often, regardless of whether they are open models or proprietary models. The proprietary models were less often hallucinating compared to the open models. The proposed statistical evaluation approach can be considered a safety sandbox where the candidate VLMs should be checked before deployment into autonomous systems as decision-making assistants and judges to analyze their own results. 

When analyzing the error categories, we identify that the overlooked traffic agents would pose a severe threat to automotive applications, as they could lead to fatal accidents and collisions where the VLM does not recognize the traffic entities that are actually present in the real environment. The statistical approach we presented allows us to identify the tasks that VLMs underperform and, henceforth, reveal the need for additional safeguarding sandboxes in automotive applications. 

While we present this approach emphasizing the need for having V\&V techniques in both design and run-time stages in the VLM safeguarding process, there are many other fields that need to be researched to improve the overall quality of VLM/LLM-assisted perception systems. Further studies are required to handle Out-of-Distribution objects or scenarios where the focused group of objects should be increased compared to the experimental setup of this study. Use of other sensor data, such as LiDAR and radar, would greatly impact this research direction. In addition to that, scaling the models and identifying smaller models with higher performance would be a promising research direction for future studies, which could help the autonomous vehicle domain. 

\section*{Acknowledgments}
This work has been supported by the Swedish Foundation for Strategic Research (SSF), grant number FUS21-0004 SAICOM, Swedish Research Council (VR) under grant agreement 2023-03810, and the Wallenberg AI, Autonomous Systems and Software Program (WASP) funded by the Knut and Alice Wallenberg Foundation.




\bibliographystyle{splncs04}
\bibliography{references}

\end{document}